\documentclass{bmvc2k}
\usepackage{booktabs}

\title{Dense and Diverse Capsule Networks: Making the Capsules Learn Better}

\addauthor{Sai Samarth R Phaye$^{*}$}{2014csb1029@iitrpr.ac.in}{1}
\addauthor{Apoorva Sikka$^{*}$}{apoorva.sikka@iitrpr.ac.in}{1}
\addauthor{Abhinav Dhall}{abhinav@iitrpr.ac.in}{1}
\addauthor{Deepti Bathula}{bathula@iitrpr.ac.in}{1}
\addinstitution{
Indian Institute of Technology Ropar\\
Punjab, India
}
\runninghead{DCNet and DCNet++}{Making the Capsules Learn Better}

\begin{document}

\maketitle

\begin{abstract}
Past few years have witnessed exponential growth of interest in deep learning methodologies with rapidly improving accuracies and reduced computational complexity. In particular, architectures using Convolutional Neural Networks (CNNs) have produced state-of-the-art performances for image classification and object recognition tasks. Recently, Capsule Networks (CapsNet) achieved significant increase in performance by addressing an inherent limitation of CNNs in encoding pose and deformation. Inspired by such advancement, we asked ourselves, can we do better?  We propose Dense Capsule Networks (DCNet) and Diverse Capsule Networks (DCNet++). The two proposed frameworks customize the CapsNet by replacing the standard convolutional layers with densely connected convolutions. This helps in incorporating feature maps learned by different layers in forming the primary capsules. DCNet, essentially adds a deeper convolution network, which leads to learning of discriminative feature maps. Additionally, DCNet++ uses a hierarchical architecture to learn capsules that represent spatial information in a fine-to-coarser manner, which makes it more efficient for learning complex data. Experiments on image classification task using benchmark datasets demonstrate the efficacy of the proposed architectures. DCNet achieves state-of-the-art performance (99.75\%) on MNIST dataset with twenty fold decrease in total training iterations, over the conventional CapsNet. Furthermore, DCNet++ performs better than CapsNet on SVHN dataset (96.90\%), and outperforms the ensemble of seven CapsNet models on CIFAR-10 by 0.31\% with seven fold decrease in number of parameters.

\end{abstract}

\section{Introduction}
\label{sec:intro} 
\begin{figure*}[t]
\begin{center}
\includegraphics[width=0.8\textwidth]{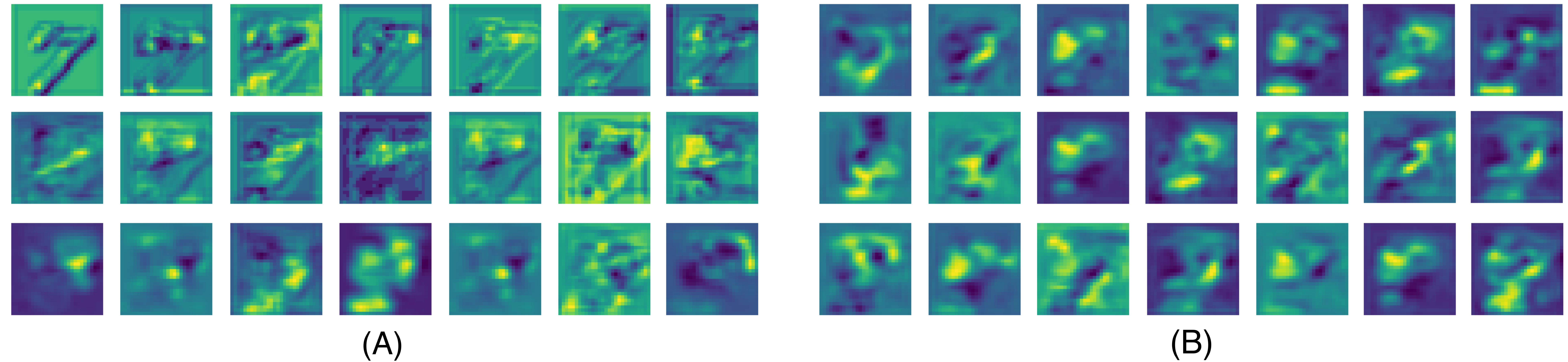}
\end{center}
   \vspace*{-5mm}
   \caption{Motivation for using dense convolutions with CapsNet: Here (A) and (B) are the feature maps at the same depth from DenseNets and CNN, respectively. It is clear that DenseNets learn clearer discriminative features as compared to CNN.}
\label{fig:visualization}
\end{figure*}

\
In recent years, deep networks have been applied to the challenging tasks of image classification \cite{Vgg, resnet, imagenet}, object recognition \cite{rec, decaf} and have shown substantial improvement. 
Many variants of CNN have been proposed by making them deeper and more complex over the time. Adding more depth in various combinations has lead to significant improvement in performance \cite{imagenet, Vgg, inception}. However, increase in depth leads to vanishing gradient problem. This issue has been addressed by ResNets \cite{resnet}, FractalNets \cite{fractalnet} by adding connections from the initial layers to the later layers. Another similar structure, which further simplifies the way skip connections are added was proposed by Huang et al.\ popularly known as the DenseNets \cite{densenet}. The network adds dense connections between every other layer in a feed-forward manner. Adding these dense connections leads to lesser number of parameters, when compared with the traditional CNN. Another benefit of concatenating these feature maps is better gradient flow across the network, which allows to train deeper networks.\\
CNNs have performed really well on various computer vision and machine learning tasks, however, it has a few drawbacks, which have been highlighted by Sabour et al.\ \cite{capsnet}. One is that CNNs are not robust to affine transformations i.e. a slight shift in the position of object make CNNs change their prediction. Although this problem can be reduced to some extent by data augmentation while training, this does not make network robust to any new pose or variation that might be present in the test data. Another major drawback is that generic CNNs do not consider spatial relationships between objects in an image while making any decision. Simply explained, CNNs only use mere presence of certain local objects in an image to make a decision while actually the spatial context of objects present is equally important. The reason is mainly the pooling operation performed in the network, which gives importance to the presence of features and ignores positional information of features, which is mainly done to decrease the\ parameters as the network grows. \\ 
To overcome these drawbacks, Sabor et al.\ \cite{capsnet} proposed a seminal architecture called the capsule networks~(CapsNet). In this model, the information is stored at the vector level instead of scalar (as in the case of the simple neural networks) and these group of neurons acting together are named as capsule. Sabour et al.\ have used the concept of routing-by-agreement and layer based squashing to achieve state-of-the-art accuracy on the MNIST dataset and detecting overlapping digits in a better way using reconstruction regularization.\\
CapsNets are really powerful, however, at the same time, there is scope for improvement in terms of complexity as the authors did not use any pooling layers and depth, as the network is currently using only one layer of convolution and capsules. On the other hand, DenseNets have the capability to achieve very high performance by feature concatenation. We compared a Dense Convolution Network having 8 layers of convolution with feature concatenation and a simple CNN having 8 layers of convolution without concatenation and 32 kernels in each layer. Both quantitative analysis (Table \ref{param}) and visualizations (Figure \ref{fig:visualization}) show that DenseNets are better at capturing diversified features. We borrow this idea of feature concatenation across layers from DenseNets \cite{densenet} as it has the potential to learn diversified features, which otherwise would require a much deeper network. We use this concept as an input to the dynamic routing algorithm of Sabour et al.\ \cite{capsnet}. \emph{The motivation behind our work is to: a) improve the performance of CapsNet, in terms of faster convergence and better results; b) achieve better performance than the CapsNet on complicated datasets such as CIFAR-10; c) try to reduce the model complexity}. 

Further, we follow the intuition behind DenseNets to design a modified decoder network with concatenated dense layers, which results in improvement in the reconstruction outputs. Our results (Table \ref{results}) are at par with state-of-art performance on the MNIST dataset \cite{MNIST} in 50 epochs, which required 1000 epochs to train conventional CapsNet. We also evaluate the proposed method on various classification datasets such as the FashionMNIST \cite{FashionMNIST}, The Street View House Numbers (SVHN) \cite{SVHN}, AffNIST Dataset\footnote{Available at: http://www.cs.toronto.edu/\textasciitilde tijmen/affNIST/} and brain tumor dataset presented in \cite{tumor, tumordataset2, capsulepapertumor}, and compare the performance of the Capsule Network and DenseCapsNet for 50 epochs keeping learning rate, learning decay rate and number of capsule parameters same. Further, we focus on why the method was not giving promising results on CIFAR-10 dataset and propose Diverse Capsule Network to improve the same.


\section{Background }
\label{sec:relatedwork}
The literature for CNN is vast. Architectures proposed using convolutions has increased significantly due to increase in computational power. The CNN tries to learn in a hierarchical manner from lower to higher layers where lower layers learn basic features like edges and higher layers learns complex features by combination of these low level features. Although, deeper networks have lead to improvement in performance, they are much more difficult to train due to huge increase in number of parameters. Recent architectures proposed, aim to improve the performance while jointly optimizing the number of parameters. Highway network \cite{highway} was the first architecture proposed in this direction to train deeper network with large number of layers. They added bypassing paths to easily train the model. ResNet \cite{resnet} model improves training by adding residual connections. Another such network proposed by Huang et. al. \cite{densenet} created a novel way of adding skip connections by introducing connections from initial convolution layers to deeper layers, naming it one dense block. 

The capsule networks \cite{capsnet} have been recently introduced to overcome the drawbacks of CNNs discussed in Section \ref{sec:intro}. Capsules are group of neurons that depict properties of various entities present in an image. There could be various properties of an image which can be captured like position, size, texture. 
Capsules use routing-by-agreement where output is sent to all final capsules. Each capsule makes a prediction for the parent capsule, which is then compared with the actual output of parent capsule. If the outputs matches, the coupling coefficient between the two capsules is increased. Let $u_i$ be an output of a capsule $i$, and $j$ be the parent capsule, the prediction is calculated as:
\begin{align*}
\hat{u}_{i|j}= \textbf{W}_{ij}u_{i}  &&&   c_{ij}= \frac{exp(b_{ij})}{\sum_k exp(b_{ik})}
\end{align*}


where $W_{ij}$ is the weighting matrix. Then coupling coefficients $c_{ij}$ are computed using a simple softmax function as shown. Here $b_{ij}$ is log probability of capsule $i$ being coupled with capsule $j$. This value is 0 when routing is started. Input vector to parent capsule $j$ is calculated as:
\vspace{-7px}
\begin{align*}
s_{j} = \sum_{i} c_{ij} \hat{u}_{j|i}  &&&   v_j = \frac{||s_j||^2}{1+||s_j||^2} \frac{s_j}{||s_j||},
\end{align*}
The output of these capsule vectors represent probability that an object represented by capsule is present in given input or not. But the output of these capsule vectors can exceed one depending on the output. Thus, a non linear squashing function defined above is used to restrict the vector length to $1$,
where $s_j$ is input to capsule $j$ and $v_j$ is output. The log probabilities are updated by computing the inner product of $v_j$ and $\hat{u_{j|i}}$. If two vectors agree, the product would be larger leading to longer vector length. In the final layer, a loss value is computed for each capsule. Loss value is high when entity is absent and capsule has high instantiation parameters. Loss is defined as:
\begin{equation*}
L_k = T_k \; max(0,\; m^+ - ||v^k||)^2 + \lambda (1 - T_k) \; max\; ( 0, ||v_k||- m^- )^2
\end{equation*}
Here $l_k$ is loss function for a capsule $k$, $T_k$ is $1$ when label is true and $0$ otherwise. 
CapsNet applies a convolution and generates $256$ feature maps. 
It is then fed to primary capsule layer where $32$, $8$D capsules are formed  using $9 \times 9$ kernel with stride $2$ followed by squashing. Finally, a DigitCaps layer is applied which forms final capsules of $16$D. 
A paper by Afshar et al.\ \cite{capsulepapertumor} shows that capsule networks are powerful enough to work on the images containing complex datasets. We aim to work on empowering capsules by using DenseNet.

\section{Our Approach}
\label{sec:approach}
We tried to customize capsule network in two frameworks which is explained in the following subsections. Furthermore, we explain the intuition behind choosing densely connected networks and then refine it to improve the performance on complex datasets.
\subsection{Dense Capsule Networks (DCNet)}
\label{sec:densecapsnet}
The feature maps learned by the first convolution layer in the baseline CapsNet model learns very basic features, which might not be enough to build capsules for complex datasets. Hence, we tried increasing the convolution layers to two and eight in the initial layer and observed that it did not lead to any improvement as shown in Table \ref{results}.  
In DCNet, we try to modify capsule networks to form a deeper architecture where we create an eight-layered dense convolutional subnetwork based on skip connections. Every layer is concatenated to the next layer in feed-forward manner, adding up to make a final convolution layer. This leads to better gradient flow as compared to directly stacked convolution layers.  

\begin{figure*}[!htbp]
\begin{center}
\includegraphics[width=0.95\textwidth]{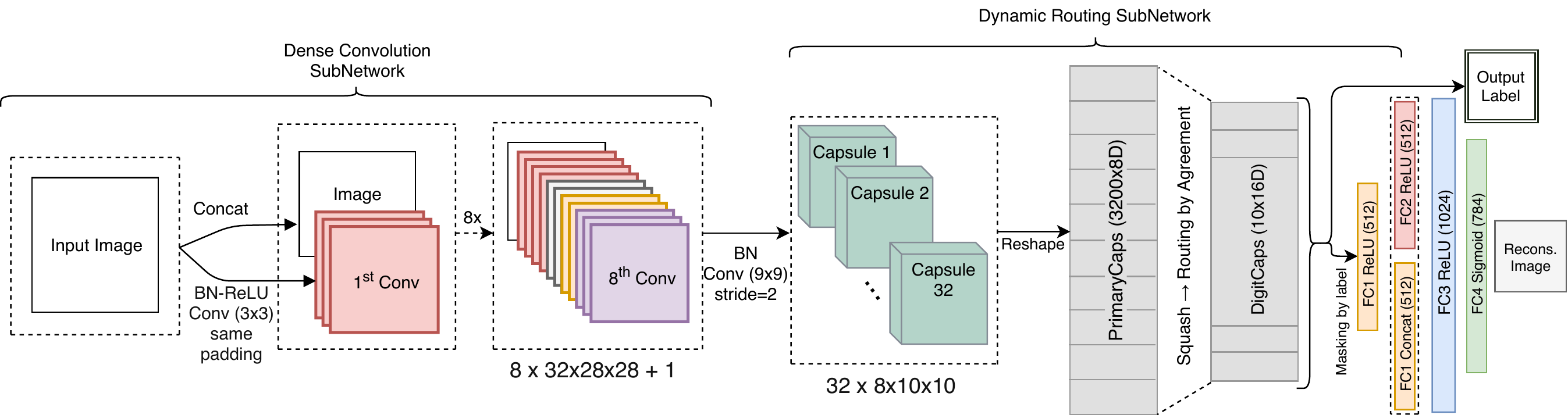}
\end{center}
   \vspace*{-4mm}
   \caption{The proposed prediction pipeline for MNIST (best viewed in color).}
\label{fig:architecture}
\end{figure*}


Figure \ref{fig:architecture} shows the detailed pipeline of the proposed architecture for MNIST dataset.
The input sample goes into 8 levels of convolutions and each of those convolution levels generates 32 new feature maps followed by concatenation with feature maps of all previous layers which results in $257$ feature maps (input image is included). These diversified feature maps act as input to the capsule layer which applies a convolution of $9 \times 9$ with a stride of $2$. The feature maps obtained act as the primary capsules of the CapsNet. The work of Sabour et al.\ \cite{capsnet} mainly focuses on equivariance instead of invariance which we totally agree with, so we did not use any of the max. or average pooling layers as used in DenseNets, which results in spatial information loss. It is important to note that while Sabour et al.\ \cite{capsnet} created the primary capsules from 256 feature maps created by the same complexity level of convolution, DCNet's primary capsules are generated by combining all the features of different levels of complexity, which further improves the classification. 

These feature maps act as thirty-two $8$D capsules, which are passed to a squash activation layer, followed by the routing algorithm. 16D final capsules are generated for each of the $10$ classes~(digits) which further generates one-hot $10$D output vector, similar to the conventional capsule network used for MNIST.

\begin{figure*}[!htbp]
\begin{center}
  \includegraphics[width=0.95\textwidth]{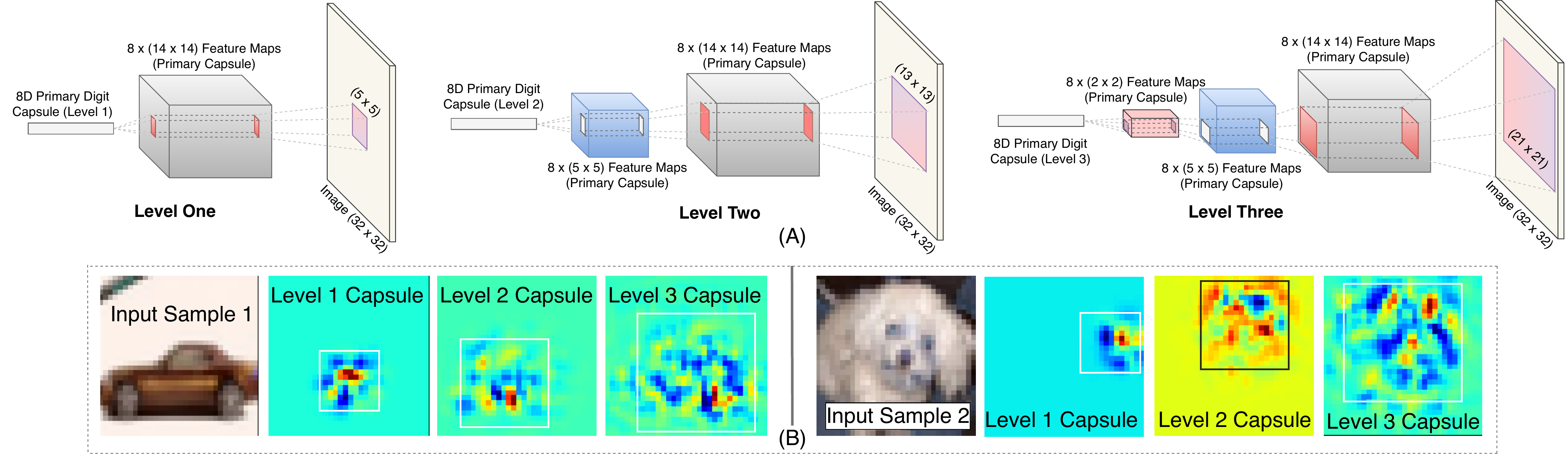}
\end{center}
\vspace*{-4mm}
\caption{Activated image region on CIFAR-10 by corresponding capsules of different levels. Row A depicts how a primary capsule gets activated on the input image. Row B shows sample activation regions on input image from capsules of different levels in DCNet++.}
\label{fig:capactivations}
\end{figure*}

Inspired by the dense connections implemented by Huang et al.\, we also modified the reconstruction model of the capsule network. The decoder is a four-layered model with concatenation of features of first layer and second layer resulting in better reconstructions, which takes the DigitCaps layer as it's input (masked by output label during training). If the size of image is more than $32 \times 32$, the number of neurons are changed from 512 to 600 and 1024 to 1200. Our experiments show significant increase in performances over various datasets like MNIST, Fashion-MNIST, SVHN. We noticed that DCNet's performance on CIFAR-10 dataset increased over single baseline CapsNet model trained with same parameters, but it did not outperform the seven ensemble model \cite{capsnet} of the capsule networks having 89.40\% accuracy. As the images in CIFAR-10 are quite complex when compared to MNIST dataset,  it is not easy for the network to encode the part-whole relationships. We address this problem by creating DCNet++ which learns well on such complex datasets.

\subsection{Diverse Capsule Networks (DCNet++)}
\label{sec:diversecapsnet}
Adding skip connections to the convolutions wasn't enough to improve performance on the CIFAR-10 dataset. This might be due to the presence of simple primary capsules which are not sufficient to encode the information present in such complex images. 
We visualized what part of an image is activated by a 8D primary capsule of DCNet by guided back-propagation and noticed that every primary capsule is generated by a small spatial area of the input image, which then act together to make a decision but these aren't enough for complex images. To overcome this, we implemented a novel way of creating primary capsules which carry information of various scales of the image, hence diversifying the capsules. This in turn helped to find connections between primary capsules of various levels of image. Figure \ref{fig:capactivations} shows which part of the image is activated in the DCNet++ model. If we consider just one level, it will be similar to image activations in DCNets and baseline CapsNet, except for the fact that the activations will have no diffusion out of the square region in Capsule Network. The activated region is diffused out in DCNet and DCNet++ because of `same' padded convolutions in the densely connected convolution layers. 
The ensemble of baseline CapsNet models (by Sabour et al.\ \cite{capsnet}) give 89.40\% accuracy on CIFAR-10 dataset, which is probably due to the fact that patches of $24 \times 24$ are used as input, which helps to activate larger spatial area for generating each primary capsule. Although it achieves reasonably higher accuracy over a single Capsule model, ensemble of seven such models leads to a huge increase in number of parameters which can be reduced for such small sized images. We focus to reduce the number of parameters used to model the data and learn better information via creating multiple levels of capsules. In the proposed model, we pass the complete image by first down sampling it to $32 \times 32$ size. A single three layered DCNet++ having 13.4M parameters achieves \textbf{89.71\%} test accuracy.

\begin{figure*}[!htbp]
\begin{center}
\includegraphics[width=0.96\textwidth]{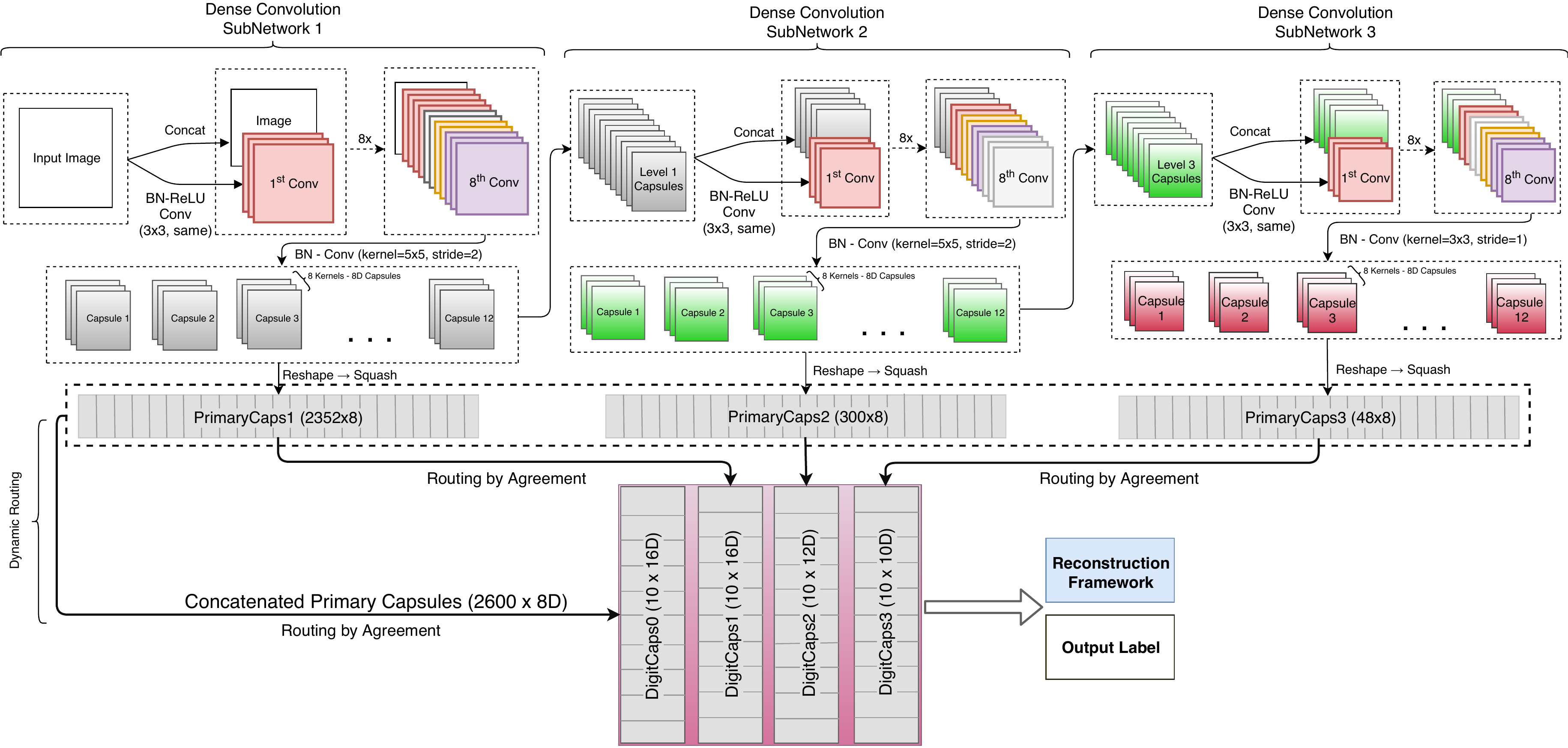}
\end{center}
   \vspace*{-4mm}
   \caption{Pipeline of three-level DCNet++ for CIFAR-10 (best viewed in color).}
\label{fig:diversecapsnet}
\end{figure*}

Figure \ref{fig:diversecapsnet} depicts the detailed pipeline of the DCNet++ model used to train the CIFAR-10 dataset. It is a hierarchical model where a DCNet model is created and it's intermediate representation is used as an input to the next DCNet which in turn generates a representation fed to the next DCNet layer. There are twelve capsules in each DCNet. A strided convolution~($9 \times 9$ size and $2$ stride) is applied which reduces the size of the image fed to next level which is similar to the concept of Pyramid of Gaussian \cite{pyramid}. This also resembles with functioning of brain which separates the information into channels. For example, there are separate pathways for high and low spatial frequency content and color information.


In addition to these three DigitCaps~(output) layers, we created one more DigitCaps output layer by routing the concatenation of three PrimaryCaps layers. The reason for adding this another level is to allow  the model to learn combined features from various levels of capsules.
We observed that in case of simple stacking of DigitCaps and joint back-propagation, the losses of last level PrimaryCaps dominate others leading to a poor learning and the former levels act as simple convolution layers. Thus, to avoid any imbalanced learning the model was jointly trained but the losses of four layers were back propagated separately.

While testing, the four DigitCaps layers are concatenated to form a 54D final capsule for each of the ten classes, and the reconstructions were created for only one channel of image using these fifty-four capsules. The reconstructions for CIFAR-10 were not very good, which we believe is due to the dominance of background noise present in the samples and presence of complex information of the image which the decoder is not robust enough to recreate. Interestingly, we notice the effect of noise in DigitCaps of different levels on the reconstruction outputs over MNIST dataset by subtracting 0.2 from each digit one at a time in the 54D DigitCaps. It is observed that the effect on reconstructions decrease from first level to the last level of capsules. DigitCaps generated from the concatenation of PrimaryCaps play a major role in affecting reconstructions, which is an additive effect of different layers of capsules (shown in Figure \ref{fig:reconstructions}).

\section{Experiments}
We demonstrate the potential of our model on multiple datasets and compare it with the CapsNet architecture. All the experiments were performed using GeForce GTX 1080 with 8GB RAM. Due to resource constraints, we ran all our models for 50 - 100 epochs. The initial learning rate was set to $0.001$ and decay rate $0.9$ with Adam as optimizer. We changed multiplier factor to scale down the reconstruction loss according to the image size so that it does not dominate margin loss. We used publicly available code \cite{densecode,capscode} for CapsNet and DenseNet to create our models\footnote{We will be sharing the code after acceptance.}. The test errors were computed once for each model. Three routings were used for all the experiments. For fair comparisons, we mostly kept the parameters of the proposed DCNet model after PrimaryCaps layer, same as the conventional CapsNet. Following are the implementation details corresponding to the datasets used:

\begin{figure*}[!htbp]
\begin{center}
\includegraphics[width=0.8\textwidth]{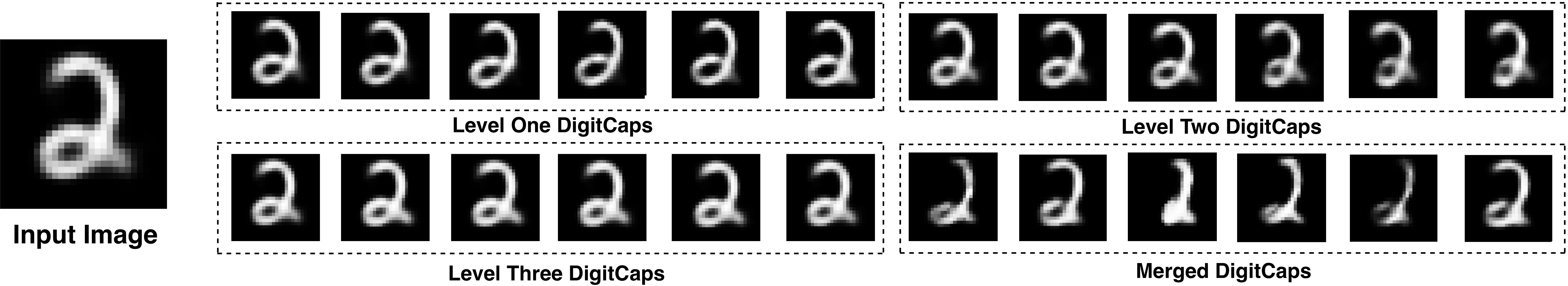}
\end{center}
\vspace{-6px}
   \caption{Reconstruction outputs by adding noise in different digits of the DigitCaps layer of DCNet++ on MNIST. Merged DigitCaps are affected the most due to the addition of all PrimaryCaps having fine-coarse activations.}
\label{fig:reconstructions}
\end{figure*}

\begin{table}[t]
\centering
{\renewcommand{\arraystretch}{01}
\resizebox{1\textwidth}{!}{%
\begin{tabular}{@{}|l|l|l|l|l|l|l|@{}}
\hline
Model                                 & MNIST                                                                                                                       & Fashion-MNIST                                                                                                        & SVHN                                       & SmallNORB                                                                                                                   & Brain Tumor Dataset                                                                                             & CIFAR-10                                                                                                                  \\ \hline
\multicolumn{1}{|l|}{\begin{tabular}[c]{@{}l@{}}Baseline\\ CapsNet\end{tabular}} & \multicolumn{1}{l|}{\begin{tabular}[c]{@{}l@{}l@{}}99.67\% (50E)\\ 99.75\% (1000E) 
\cite{capsnet}\\ 99.65\% (1000E, NR) 
\cite{capsnet}\end{tabular}}                    & \multicolumn{1}{l|}{93.65\% (100E)}                                                                            & \multicolumn{1}{l|}{\begin{tabular}[c]{@{}l@{}}93.23\% (100E) \\ 95.7\% ($>$ 100E) \cite{capsnet}
\end{tabular}} & \multicolumn{1}{l|}{89.56\% (50E)}                                                                                    & \multicolumn{1}{l|}{\begin{tabular}[c]{@{}l@{}}78\% (10E)\cite{capsulepapertumor}\\ 87.5\% (50E)\end{tabular}}              & \multicolumn{1}{l|}{\begin{tabular}[c]{@{}l@{}}89.40\%  \cite{capsnet}\\ (7 model \\ensemble)\end{tabular}}                                                                            \\ \hline
\multicolumn{1}{|l|}{DCNet}          & \multicolumn{1}{l|}{\textbf{\begin{tabular}[c]{@{}l@{}}99.75\% (50E)\\ 99.71\% (50E, NR)\\ 99.72\% (50E, BR)\end{tabular}}} & \multicolumn{1}{l|}{\begin{tabular}[c]{@{}l@{}}94.64\% (100E)\\ 94.59\% (100E, NR)\end{tabular}} & \multicolumn{1}{l|}{95.58\% (50E)}   & \multicolumn{1}{|l|}{94.43\% (50E)}                                                                                    & \multicolumn{1}{l|}{\begin{tabular}[c]{@{}l@{}}86.8\% (10E)\\ 93.04\% (50E)\end{tabular}}           & \multicolumn{1}{l|}{\begin{tabular}[c]{@{}l@{}}82.63\%\\ (single model)\end{tabular}}                                                                                   \\ \hline
\multicolumn{1}{|l|}{DCNet++}        & \multicolumn{1}{l|}{99.71\% (50E)}                                                                                                     & \multicolumn{1}{l|}{\textbf{94.65\% (100E)}}                                                                                               & \multicolumn{1}{l|}{\textbf{96.90\% (50E)}}                    & \multicolumn{1}{l|}{\textbf{\begin{tabular}[c]{@{}l@{}}95.34\% (75E)\\ 95.27\% (75E, NR)\end{tabular}}} & \multicolumn{1}{l|}{\begin{tabular}[c]{@{}l@{}}71.43\% (10E)\\ \textbf{95.03\% (50E)}\end{tabular}} & \multicolumn{1}{l|}{\textbf{\begin{tabular}[c]{@{}l@{}}89.71\% (120E)\\ 89.32\% \\(120E, NR)\end{tabular}}} \\\hline
\end{tabular}%
}
}
\vspace{-5px}
\caption{Performance of CapsNets, DCNets and DCNet++ over various datasets. NR - No Reconstruction SubNetwork, E - training epochs, BR - baseline Reconstruction SubNetwork.}
\label{results}
\end{table}

 \textbf{MNIST Handwritten digits dataset} and \textbf{Fashion-MNIST dataset} have $60$K and $10$K training and testing images with each image size being $28 \times 28$ in size. We did not use any data augmentation scheme and repeated the experiment $3$ times. It can be clearly seen in Table \ref{results} that DCNet is able to learn the input data variations quickly, as compared to the CapsNet, i.e., it is able to achieve 99.75\% test accuracy on MNIST and 94.64\% on Fashion-MNIST dataset with 20 fold decrease in total iterations. We changed the stride from two to one in the PrimaryCaps layer of second level of DCNet++ to fit the image. We did not observe any improvement in the  performance of 3-level DCNet++ on MNIST dataset which is as expected because we are capturing fine-to-coarse level features, due to which every tiny variation in the writing of a particular number will be captured from the training set. These `fine' features might be causing a problem during the testing phase. We need to invest more time into improving it.

\textbf{CIFAR-10 dataset} is an image dataset having 50K and 10K samples for training and testing. We used channel means and standard deviations for normalizing the data. We compare our proposed model with the ensemble of four and seven capsule network models \cite{capsnet} for first 50 epochs. We used similar parameters for DCNet model that were used for MNIST and the test accuracy improved over the ensemble of four CapsNet models (implemented by Xi et al.\ \cite{complexdata}) to $82.68\%$ from  $71.55\%$. The proposed DCNet++ model resulted in \textbf{89.7\%} accuracy in $13.4$M parameters, which is significantly less than $7$ ensemble model proposed by Sabour et al.\ ($7 \times 14.5$M, $89.40\%$)  and DCNet (16M parameters for $32 \times 32$ image).
  
\begin{table}[!htbp]
\centering
{\renewcommand{\arraystretch}{1}
\resizebox{0.85\textwidth}{!}{%
\begin{tabular}{@{}|l|l|l|l|@{}}
\hline
\textbf{Model}                                             & \textbf{Description}                                                                                                                                                      & \textbf{Parameters}                 & \textbf{Test Acc.} \\ \hline
\multicolumn{1}{|l|}{CapsNet (Baseline)}    & \multicolumn{1}{l|}{Conv - Primary Capsules - Final Capsules}                                                                                                                                           & \multicolumn{1}{l|}{8.2M}  & \multicolumn{1}{l|}{99.67\%}          \\ \hline
\multicolumn{1}{|l|}{CapsNet Variant} & \multicolumn{1}{l|}{\begin{tabular}[c]{@{}l@{}}Added one more initial convolution layer having\\ 256 9x9 kernels with same padding, stride=1\end{tabular}} & \multicolumn{1}{|l|}{13.5M} & \multicolumn{1}{l|}{99.66\%}          \\ \hline
\multicolumn{1}{|l|}{DCNet Variant One}    & \multicolumn{1}{l|}{\begin{tabular}[c]{@{}l@{}}Used 3 convolution layers with 8 feature maps each\end{tabular}}                                                & \multicolumn{1}{|l|}{6.9M}  & \multicolumn{1}{l|}{99.66\%}          \\ \hline
\multicolumn{1}{|l|}{DCNet Variant Two}    & \multicolumn{1}{l|}{\begin{tabular}[c]{@{}l@{}}Removed the concatenations (no skip connections)\\ acting like simple 8 layered convolutions\end{tabular}}                                                                                                                  & \multicolumn{1}{|l|}{4M}    & \multicolumn{1}{l|}{99.68\%}          \\ \hline
\multicolumn{1}{|l|}{DCNet Variant Three}    & \multicolumn{1}{l|}{\begin{tabular}[c]{@{}l@{}}Used 8th convolution layer only (no concatenation\\ in the last layer) to create Primary Capsules\end{tabular}}                       & \multicolumn{1}{l|}{7.2M}  & \multicolumn{1}{l|}{\textbf{99.72\%}} \\ \hline
\multicolumn{1}{|l|}{Final DCNet}       & \multicolumn{1}{l|}{DenseConv - Primary Capsules - Final Capsules}                                                                                                                                           & \multicolumn{1}{l|}{11.8M} & \multicolumn{1}{l|}{\textbf{99.75\%}} \\ \hline
\end{tabular}
}
}
\vspace{1mm}
\caption{Comparison of various model variations after 50 epochs on MNIST dataset. For details of proposed DCNet, refer to Section \ref{sec:densecapsnet}.}
\label{param}
\end{table}

 \textbf{Street View House Numbers}~(SVHN) contains 73K and 26K real-life digit images for training and testing. To compare results with the original capsule network model \cite{capsnet} for first 50 epochs, we modified DCNet by using 6D 16 primary capsules with 8D final capsule from four convolution layers with 18 feature maps. The results of the DCNet model improved over the replicated CapsNet model to test accuracy of \textbf{95.59\%} from 93.23\%. The model for DCNet++ is same as what we used for CIFAR-10 dataset, resulting in \textbf{96.9\%} accuracy, compared to 95.70\% by CapsNets (more than 50 epochs) \cite{capsnet}.
 
 \textbf{Brain Tumor Dataset} contains 3,064 MRI images of 233 patients diagnosed with one of the three brain tumor types (i.e., meningioma, glioma, and pituitary tumor). Afshar et al.\ \cite{capsulepapertumor} changed the capsule network architecture with initial convolution layer having $64$ feature maps. We created an equivalent DCNet model~(for uniform results) by modifying the eight initial convolution layers to four layers with $16$ kernels each, totaling to $64$ kernels and decreased the primary capsules to $6$. We also changed the learning rate of our model to $1E-4$ for effective learning. The DCNet++ is a 3-level hierarchical model of the modified DCNet. We trained the models on eight fold cross validation (results in Table \ref{results}).

 \textbf{SmallNORB dataset} has $24$K and $24$K images for training and testing. We normalized the images and add random noise and contrast to each image. The proposed model was compared with the CapsNet \cite{capsnet} for first 50 epochs, by modifying DCNet taking $6$D $16$ primary capsules with $8$D final capsule from four convolution layers with $18$ kernels each. The results of DCNet improved over the replicated CapsNet model to give a test accuracy of $\textbf{95.58\%}$. The DCNet++ model is same as what we used for CIFAR-10 dataset, resulting in \textbf{96.90\%} accuracy, compared to CapsNet \cite{capsnet}, which achieved $95.70\%$ in more than 50 epochs.
 
\begin{figure}[t]
   \begin{minipage}{0.48\textwidth}
     \centering
		\includegraphics[width=0.95\textwidth]{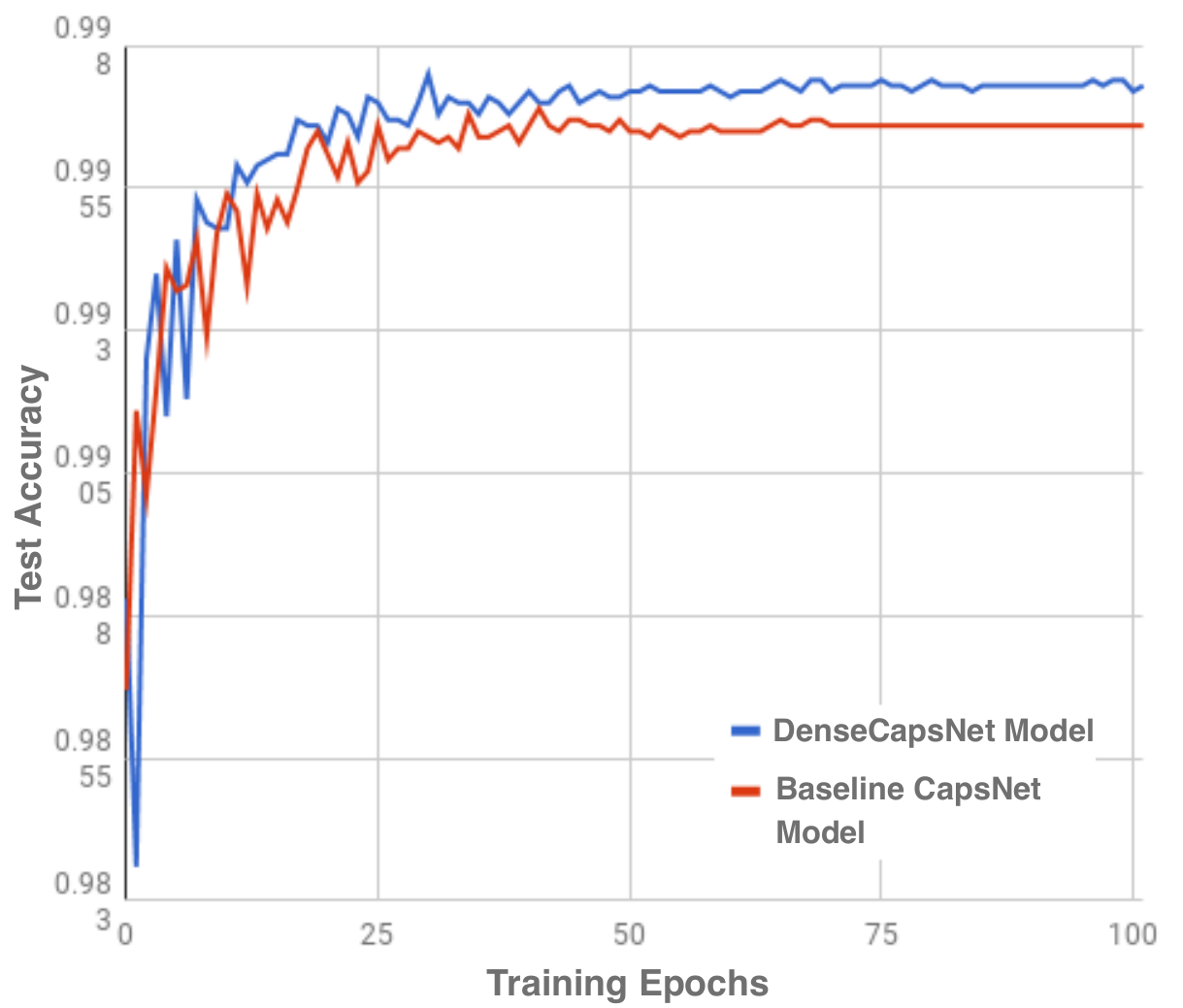}
        \vspace{3px}
   		\caption{Comparison of the performance of DCNet and the CapsNet model on MNIST dataset.}
		\label{fig:accuracy}
   \end{minipage}\hfill
   \begin {minipage}{0.48\textwidth}
     \centering
		\includegraphics[width=0.95\textwidth]{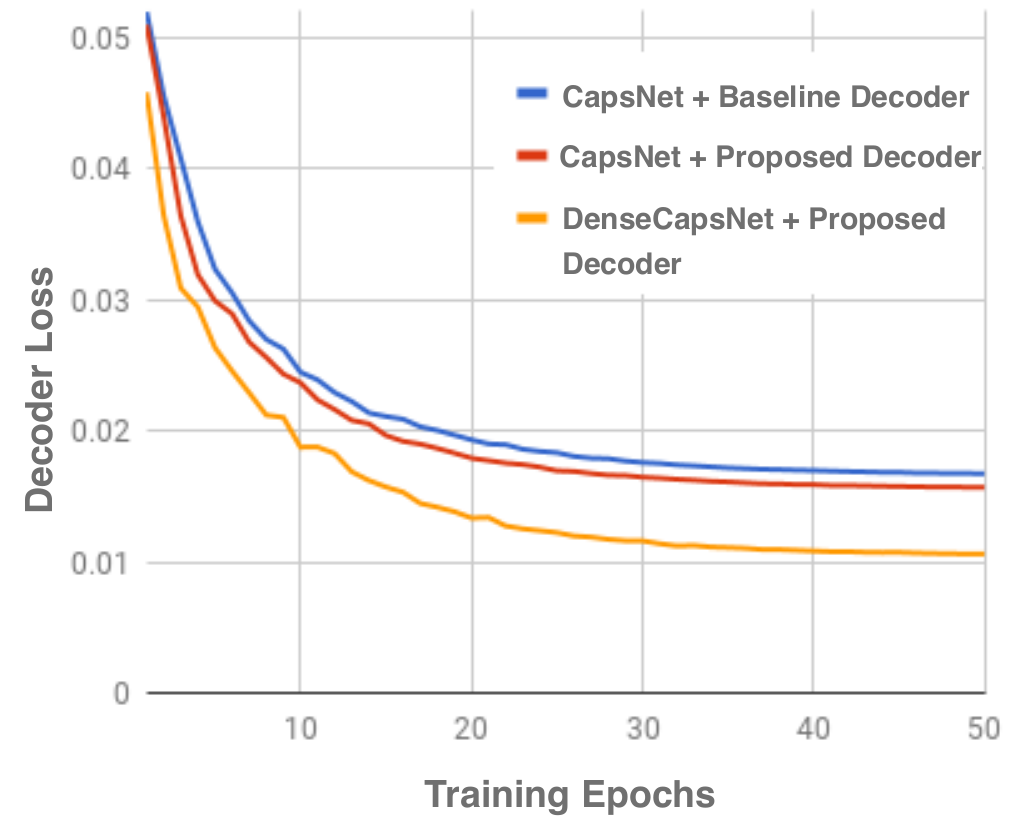}
		\label{fig:reconstructionLoss}
        \vspace{-8px}
   		\caption{Reconstruction MSE loss of CapsNet+Baseline Decoder, CapsNet+Proposed Decoder and DCNet+Proposed Decoder on MNIST dataset.}
   \end{minipage}  
\end{figure}

We find that the loss of the modified decoder in the DCNet decrease by a considerable amount (shown in Figure \ref{fig:reconstructionLoss}), having same loss multiplier factor as that of CapsNet. Also, Figure \ref{fig:accuracy} shows the test accuracy comparison of CapsNet and DCNet from which we can clearly infer that DCNet has a faster convergence rate.

\section{Conclusion}
We proposed a modification of Capsule Network, DCNet that replaces standard convolution layers in CapsNet with densely connected convolution. Addition of direct connections between two consecutive layers help learn better feature maps, which in turn helps in forming better quality primary capsules. The effectiveness of this architecture is demonstrated by state-of-the-art performance (99.75\%) on MNIST data with twenty fold decrease in total training iterations, over conventional CapsNets. Although the same DCNet model performed better (82.63\%) than a single, baseline CapsNet model on CIFAR-10 data, it was below par compared to seven ensemble model of CapsNet with 89.40\% accuracy.

Performance of CapsNet on real-life, complex data (CIFAR-10, ImageNet, etc.) is known to be substandard compared to simpler datasets like MNIST. DCNet++, addresses this limitation by stacking multiple layers of DCNet to enhance the representational power of the network. The hierarchical structure helps learn intricate relationships between fine-to-coarse level features. A single, three-level DCNet++ achieves 89.71\% accuracy on CIFAR-10, which is an increase of 0.31\% accuracy over seven ensemble CapsNet model with significantly less number of parameters. The proposed networks substantially improve the performance of existing Capsule Networks on other datasets such as SVHN, SmallNORB and Tumor Dataset. In future, we plan to incorporate EM routing \cite{EM} in DCNet and DCNet++ and work on reducing the computational complexity of the model even further.

\bibliography{egbib}
\end{document}